\documentclass[sigconf]{acmart}

\usepackage{graphicx}
\usepackage{amsmath}
\usepackage{multirow}
\usepackage{tabularx}
\usepackage{colortbl}
\usepackage{subcaption}
\usepackage{diagbox}
\usepackage{amsfonts}
\usepackage{enumitem}

\usepackage{booktabs}
\usepackage{caption}

\AtBeginDocument{%
  \providecommand\BibTeX{{%
    \normalfont B\kern-0.5em{\scshape i\kern-0.25em b}\kern-0.8em\TeX}}}

\begin{document}


\title[EcoSense]{EcoSense: Energy-Efficient Intelligent Sensing for In-Shore Ship Detection through Edge-Cloud Collaboration}


\author{Wenjun~Huang}
\affiliation{%
  \institution{University of California, Irvine}
  \city{Irvine}
  \state{California}
  \country{USA}
}

\author{Hanning Chen}
\affiliation{%
  \institution{University of California, Irvine}
  \city{Irvine}
  \state{California}
  \country{USA}
}

\author{Yang Ni}
\affiliation{%
  \institution{University of California, Irvine}
  \city{Irvine}
  \state{California}
  \country{USA}
}

\author{Arghavan Rezvani}
\affiliation{%
  \institution{University of California, Irvine}
  \city{Irvine}
  \state{California}
  \country{USA}
}

\author{Sanggeon~Yun}
\affiliation{%
  \institution{University of California, Irvine}
  \city{Irvine}
  \state{California}
  \country{USA}
}

\author{Sungheon~Jeon}
\affiliation{%
  \institution{University of California, Irvine}
  \city{Irvine}
  \state{California}
  \country{USA}
}

\author{Eric~Pedley}
\affiliation{%
  \institution{University of California, Irvine}
  \city{Irvine}
  \state{California}
  \country{USA}
}

\author{Mohsen~Imani}
\affiliation{%
  \institution{University of California, Irvine}
  \city{Irvine}
  \state{California}
  \country{USA}
}








\renewcommand{\shortauthors}{W. Huang, et al.}

\begin{abstract}
Detecting marine objects inshore presents challenges owing to algorithmic intricacies and complexities in system deployment.
We propose a difficulty-aware edge-cloud collaborative sensing system that splits the task into object localization and fine-grained classification. 
Objects are classified either at the edge or within the cloud, based on their estimated difficulty.
The framework comprises a low-power device-tailored front-end model for object localization, classification, and difficulty estimation, along with a transformer-graph convolutional network-based back-end model for fine-grained classification.
Our system demonstrates superior performance (mAP@0.5 \textbf{+4.3\%}) on widely used marine object detection datasets, significantly reducing both data transmission volume (by \textbf{95.43\%}) and energy consumption (by \textbf{72.7\%}) at the system level. We validate the proposed system across various embedded system platforms and in real-world scenarios involving drone deployment.
\end{abstract}

\maketitle

\section{Introduction}
Inshore Marine Object Detection (IMOD) constitutes a crucial undertaking with diverse applications, including but not limited to vessel identification and positioning, collision avoidance systems, safe autonomous navigation, and search and rescue missions \cite{zhang2021survey}. 
This task primarily involves the analysis of data originating from a range of sensors, such as radio detection and ranging (RADAR), automatic identification system (AIS), infrared cameras, and RGB cameras. 
Currently, widely used methods mostly employ RADAR and AIS sensors \cite{yu2021object}. 
However, these methodologies exhibit inherent deficiencies that compromise their applicability. 
Specifically, AIS is contingent upon all vessels being equipped with AIS devices and actively transmitting signals, presenting a substantial challenge in detecting entities without AIS capabilities and addressing potential misuse issues. 
Conversely, RADAR technology is limited by its susceptibility to blind spots and its incapacity to discern detailed physical characteristics, thereby restricting its ability to provide exhaustive information critical for safety assessments. 
Similarly, the utility of infrared cameras is curtailed by their limited detection range and the prohibitive costs associated with models offering sufficient range for civilian applications.
As a result, RGB cameras are gaining increasing prominence among these sensors due to their rich information, cost-effectiveness, and widespread accessibility.

With the progress in computer hardware and associated tools, deep learning has gained widespread adoption in the field of RGB camera-based object detection \cite{ren2015faster, redmon2016you}. 
However, these conventional object detectors cannot be applied directly to IMOD due to various challenges, one of which arises from the limited inter-class variation observed among the subordinate categories of ships. 
Even though different ships have unique functionalities, their relatively uniform shapes and appearances pose difficulty in distinguishing between them. 
This inherent characteristic renders IMOD a challenging target for machine learning algorithms.

\begin{figure*}[!t]%
\centering
\includegraphics[width=1\textwidth]{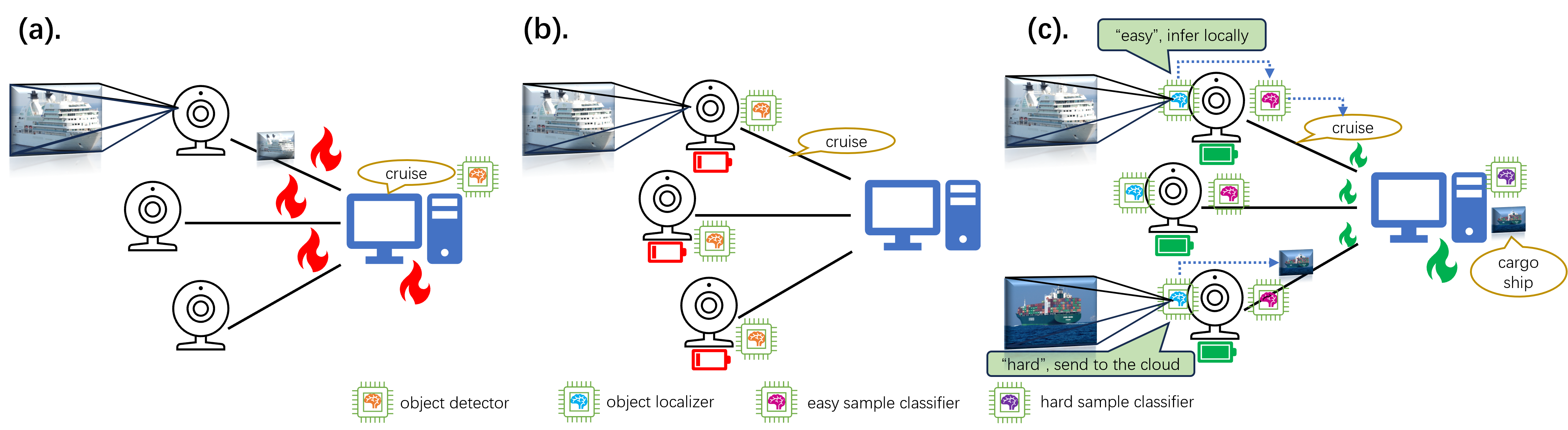}
\caption{\textbf{Comparison of different framework design}. \textbf{(a). Centralized manner.} The model is deployed on the cloud, leading to high pressure on bandwidth and the cloud. \textbf{(b). Distributed manner.} The model is deployed near each sensor. The inference is implemented locally and only the result is sent out. \textbf{(c). Our edge-cloud collaboration framework. }A powerful classifier is deployed on the cloud. Each edge device has an object localizer and a lightweight classifier. Depending on the difficulty of the ROI, the detected objects are classified on the edge or transmitted to the cloud for classification. The framework makes a good balance between edge device battery life, accuracy, and data transmission volume.}\label{fig: motivation}
\end{figure*}

Apart from algorithmic challenges, IMOD also suffers from difficulties in real-world deployments.
Sensors for IMOD, such as coastal webcams, drones, and movable cameras, are deployed in vast numbers and in challenging environments along lengthy and narrow coastlines. 
Such an elevated complexity of the IMOD systems requires the awareness of platform and environmental constraints and optimized utilization of computational resources.
In addition to the techniques of image compression \cite{zhang2023ultra, mao2022trace} and memory management for edge devices \cite{wen2020hardware, wen2021openmem, wen2022software, wen2021fpga}, 
previous efforts to tackle this challenge can generally be categorized into two approaches: centralized and distributed, as illustrated in Fig.~\ref{fig: motivation}.
While both approaches alleviate the problem to some extent, they exhibit intrinsic drawbacks. 
In a centralized approach (Fig.~\ref{fig: motivation}(a)), the detection model is deployed on a central powerful server (i.e., cloud), and all sensors transmit data to the server for inference. 
Despite delivering outstanding performance, this approach escalates the volume of data transferred over the wireless channel, demanding high bandwidth and reliability. 
Furthermore, any anomaly occurring on the server can lead to a system-wide shutdown.
Conversely, a distributed approach, as presented in Fig.~\ref{fig: motivation}(b) and explored in \cite{sharafaldeen2022marine, liu2024enhanced}, involves deploying the detection model in proximity to each sensor. 
In this configuration, the inference process occurs near the sensor, with only the results transmitted externally. 
While this approach conserves bandwidth and enhances resilience against potential system failures, it grapples with challenges related to battery life, execution time, and overall performance owing to limitations imposed by edge devices.

Acknowledging the constraints inherent in the aforementioned architectures, our proposed solution entails an edge-cloud collaboration framework incorporating a customized hierarchical model. 
The objective is to strike a balance between energy consumption, accuracy, and speed. The contributions of this work are delineated as follows:
\begin{itemize}[leftmargin=*]
    \item To tackle the algorithmic challenge, we proposed an edge-cloud collaboration framework that splits IMOD into object localization and fine-grained classification (FGC). 
    \item We developed a lightweight model for object localization, difficulty estimation, and classification, for low-power edge devices. 
    \item We developed a transformer-based model combined with a graph neural network (GNN) specializing in FGC.
    \item The system was deployed across platforms and tested in real-world scenarios, specifically employing a drone for evaluation.
    \item Experiments encompassing the entire system demonstrate that our system attains commendable accuracy while significantly reducing both the system's energy consumption and the volume of transmitted data.
\end{itemize}

\section{Related Work}
\subsection{Fine-grained Image Analysis}
Fine-grained image analysis (FGIA) is a longstanding problem and underpins a diverse set of real-world applications \cite{wei2021fine}. 
It targets analyzing visual objects from subordinate categories, e.g., species of birds
, models of cars
, ships \cite{lee2018image, shao2018seaships}, etc. The small inter-class and large intra-class variation makes it a challenging problem. 

Researchers have endeavored to develop models capable of identifying the distinctive semantic components of fine-grained objects and subsequently constructing an intermediate representation aligning with these components for ultimate classification purposes. \cite{zhang2016weakly} utilized a spatial pyramid strategy to generate part proposals derived from object proposals. Subsequently, through a clustering approach, they established prototype clusters for part proposals and further refined them to obtain discriminative part-level features. Cosegmentation-based methods, as presented by \cite{guillaumin2014imagenet}, are also commonly used in FGIA. One approach involves leveraging cosegmentation to derive object masks in the absence of supervision, followed by heuristic strategies such as part constraints \cite{he2017weakly} to delineate fine-grained parts. It is pertinent to note that prior research predominantly overlooks the internal semantic interrelations among discriminative part-level features. Specifically, existing methodologies select discriminative regions in isolation and directly utilize their features, thus disregarding the mutual semantic correlations among an object's features and the potential discriminative power of region groups.

\subsection{Object Detection}
Object detection is one branch of computer vision widely applied in people’s lives, such as monitoring security \cite{joshi2012survey}.
The majority of CNN-based object detection models can generally be categorized into two classes: two-stage detectors \cite{girshick2015fast} and one-stage detectors \cite{ redmon2016you}.
Two-stage detectors such as Faster R-CNN~\cite{ren2015faster} first suggest candidate object bounding boxes via a Region of Interest (RoI) pooling layer, and the second stage involves classification and refinement processes upon features extracted from bounding boxes.
In contrast, one-stage detectors forego the region proposal and directly perform detection over a dense sampling of locations \cite{redmon2016you}.
On the other hand, the emergence of Transformer architecture has demonstrated its superiority over conventional CNN methods in certain tasks \cite{liu2021swin}. 
\cite{yu2024credit} has shown the effectiveness of combining Transformer models with T-SNE to build efficient detection models, highlighting the critical role of these models in improving detection accuracy and efficiency by capturing complex data patterns.
Nevertheless, the deployment of these models on resource-constrained devices is hindered due to their substantial computational demands. 
Although efforts have been made to develop lightweight variants, such endeavors often lead to compromised performance \cite{han2020ghostnet, hu2018squeeze}.

\section{Method}

\begin{figure*}[!t]%
\centering
\includegraphics[width=1\textwidth]{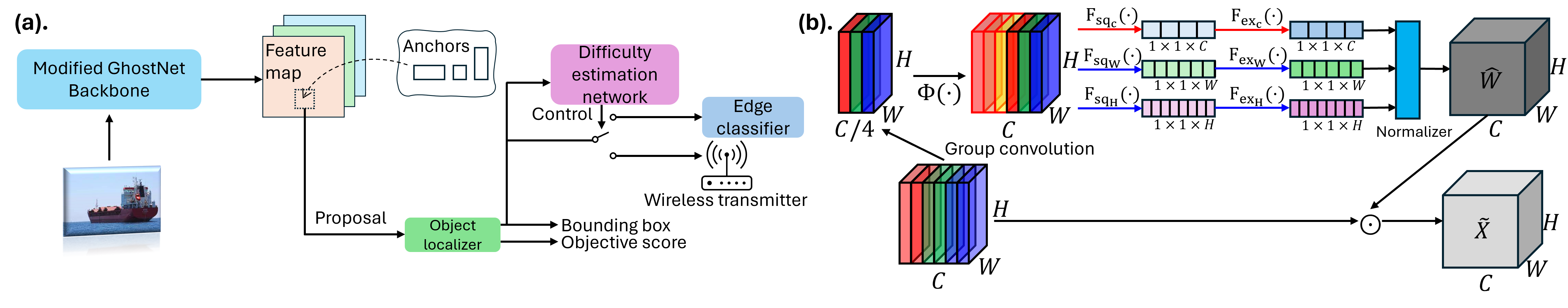}
\caption{\textbf{Front-end model} \textbf{(a). Schematic of front-end model} \textbf{(b). Modified Ghost module}. $\Phi(\cdot)$ denotes the linear projection in the original Ghost module. Red arrows show the path of the Squeeze-Excitation operation, and blue arrows show Coordinate Attention ($F_{sq}(\cdot)$ denotes squeeze operation, $F_{ex}(\cdot)$ denotes excitation operation). $\odot$ denotes element-wise multiplication.}
\label{fig: all attention schematic}
\end{figure*}

Inspired by the principles underlying two-stage object detectors and the concept of disentangling recognition and localization \cite{singh2018r, yun2024hypersense, huang2024plug}, our proposed divides the task into precise object localization and classification, as depicted in Fig.~\ref{fig: motivation}(c). 
Object localization occurs on the edge devices. Following localization, the generated proposals are directed to different classifiers based on their difficulty estimated by a difficulty estimator.
The object localizer, the difficulty estimator, along with an ``easy sample" classifier deployed on edge devices, collectively referred to as \textit{front-end model}, aims to minimize the need for data transmission. In addition, a ``hard sample" classifier, referred to as \textit{back-end model}, is deployed on the cloud to ensure the provision of reliable predictions. 
This framework strikes a well-balanced compromise between performance and energy efficiency.

\subsection{Front-end Model}

The schematic of our front-end model is illustrated in Fig.~\ref{fig: all attention schematic}(a). 
Considering constraints related to model size and energy efficiency, our approach employs a modified Faster R-CNN with an improved
GhostNet \cite{han2020ghostnet}
as the backbone. 
Fig.~\ref{fig: all attention schematic}(b) illustrates our modified Ghost module.
The ghost module generates more ghost feature maps from a set of low-cost linear projections ($\Phi(\cdot)$), revealing information underlying the intrinsic features. 

On  the top of ghost module, we introduce attention modules based on squeeze-and-excitation in MobileNetV3
\cite{hu2018squeeze}.
The \textit{squeeze} module, denoted as $F_{sq_c}(\cdot)$, performs channel-wise compression of descriptors through global average pooling, aiming to leverage channel dependencies. 
The \textit{excitation} module ($F_{ex_c}(\cdot)$) is implemented using two fully connected layers, followed by a sigmoid activation function, and is connected to the \textit{squeeze} operator. 
\textit{Excitation} maps the input-specific descriptor to a set of channel weights. 

However, the squeeze-and-excitation mechanism overlooks the influence of positional information, which holds significance in the context of object localization and detection. Recognizing this challenge, we introduced the concept of ``coordinate attention'' \cite{hou2021coordinate}. In contrast to channel attention, which transforms a feature tensor into a single feature vector (with dimension $\mathbb{R}^{1\times 1\times C}$) through 2D global pooling, coordinate attention decomposes attention into two 1D feature encoding processes along the two spatial directions, respectively.
This approach allows for the capture of long-range dependencies along one spatial direction, while simultaneously preserving precise positional information along the other spatial direction. The resulting descriptors 
can be complementarily applied to the input feature map to enhance the representations of objects. 

Within our design, attention modules operate independently on a single input in parallel. The calculated 1D features from the excitation operators in three dimensions are channeled into a normalizer. The output from the normalizer ($\hat{W}$) maintains the same shape as the input ($X$), with a summation equal to 1. The final output ($\tilde{X}$) is obtained through element-wise multiplication ($\odot$) between $X$ and $\hat{W}$. This design choice facilitates parallelized attention processing and enhances the model's ability to capture both channel-wise and spatial information effectively.
To keep the dimensionality consistent, we use group convolution \cite{krizhevsky2012imagenet} before the linear projection.
It results in multiple channel outputs per layer, leading to wider networks and helping a network learn a varied set of features.


The object localizer adopts a Region Proposal Network (RPN) followed by a few fully connected layers refining the bounding boxes.
It generates bounding boxes for each detected object and outputs the object score.
To mitigate redundant generation, we incorporate non-maximum suppression (NMS). 
This NMS step effectively filters out repetitive proposals related to the same object, resulting in the generation of a refined proposal that encapsulates the entire shape of the object.
The loss function of the localizer has two terms:
\begin{equation}
    L_{localizer}=l_{obj}+l_{reg}
\end{equation}
$l_{obj}$ represents the loss of objectness scores, and $l_{reg}$ represents the loss of bounding box regression offset.
The ``easy sample'' classifier is a fully convolutional network that follows the settings in \cite{singh2018r}.

During the inference, the model assesses the difficulty of the proposals and subsequently decides whether to classify them on the edge or send them to the back-end model.
We introduce a dedicated ``difficulty estimation network'', comprising two fully connected layers, after the localizer. 
The input to this branch is the proposal itself, and the output is the estimated difficulty associated with that specific proposal. 
The difficulty estimation network is trained after fixing the parameters of the front-end model. The details are elaborated in \ref{subsection: dataset}.


\begin{figure}[!t]%
\centering
\includegraphics[width=1\columnwidth]{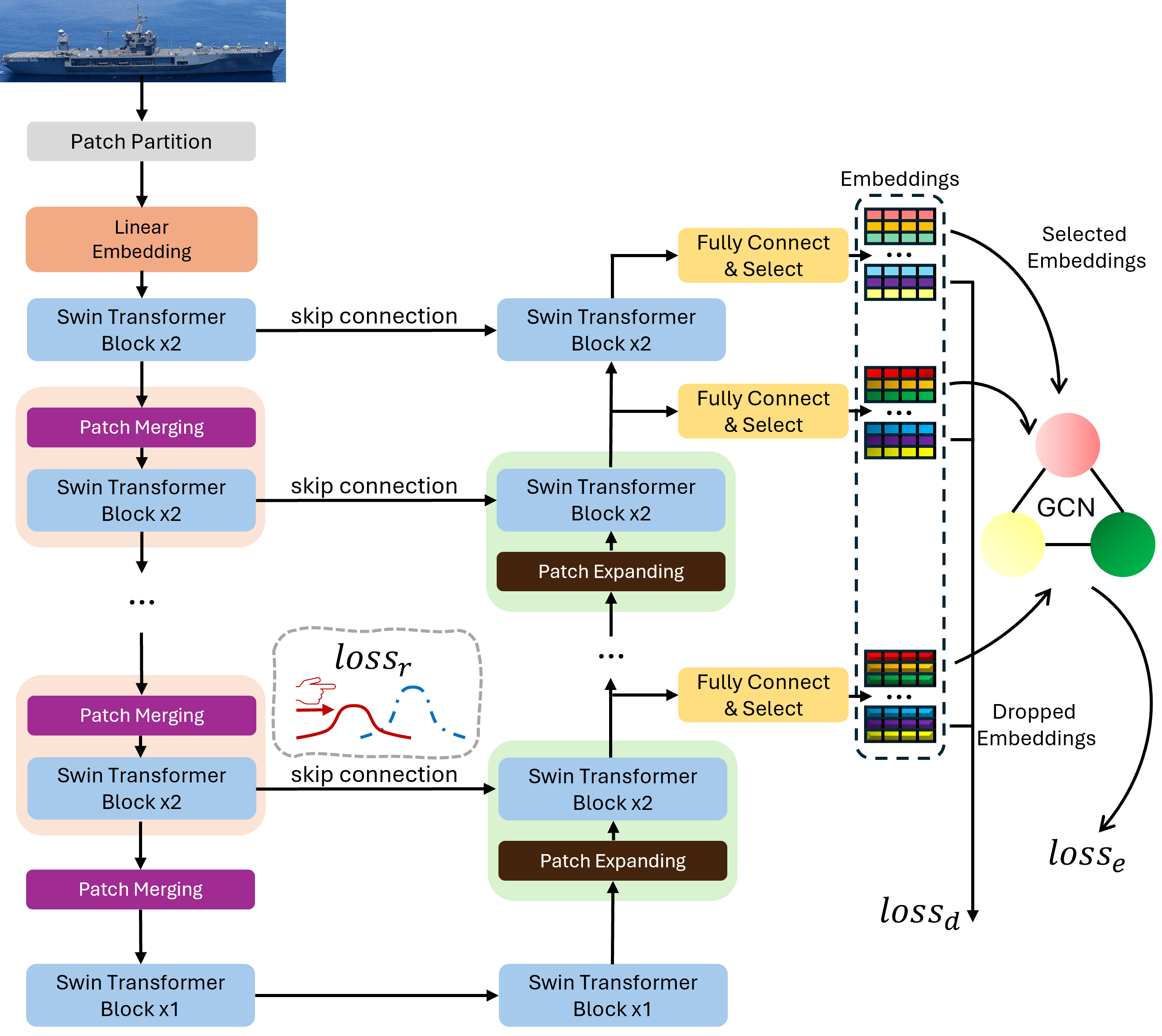}
\caption{\textbf{Schematic of attention module}. $F_{sq}(\cdot)$ denotes squeeze operation, $F_{ex}(\cdot)$ denotes excitation operation, $\odot$ denotes element-wise multiplication.}
\label{fig: back-end schematic}
\end{figure}

\subsection{Back-end Model}
Given the high similarity between categories, we leverage a robust Swin-Transformer \cite{liu2021swin} combined with a Graph Convolutional Network (GCN) for improved classification of hard samples, inspired by the approach presented in \cite{jia2023exploiting}. 
In addition, we follow \cite{zheng2024advanced}, which utilizes the Synthetic Minority Over-sampling Technique (SMOTE) combined with XGBoost to improve the robustness.
A schematic of our back-end model is depicted in Fig.~\ref{fig: back-end schematic}.
However, prior models often concentrated on localizing subtle discrepancies while neglecting the valuable information provided by the background. 
This information can indicate which features are unnecessary or potentially harmful for accurate classification.
To address the limitations, we introduced a Background Suppression (BS) module aimed at mitigating unnecessary background interference while augmenting the representation of diverse features \cite{chou2023fine}.
\subsubsection{Background Suppression}
Given a feature map $h_i\in \mathbb{R}^{C_i\times H_i\times W_i}$, we calculate the embedding map of each category 
\begin{equation}
    Y_i = W_i\cdot h_i + b_i
\end{equation}
with dimensions $\mathbb{R}^{C_{gt}\times (H_i \times W_i)}$ using a fully-connected layer, where $i$ denotes the $i$-th block, $C_{gt}$ denotes the number of categories.
Then, we calculate the score map 
\begin{equation}
    P_i = \text{softmax}(Y_i)
\end{equation}
and select the embeddings with the top-K score, denoted by $Y_{s_i}$.
These selected embeddings from all blocks ($Y_{s_i}, i\in {1,\dots,n}$) are concatenated to create a graph vertex embedding $Y_e$, which is then input into a GCN to capture the relationships between categories and facilitate predictions. The unselected embeddings are concatenated to generate a dropped embedding matrix $Y_{d}$. 
The loss function for classification is a standard cross-entropy function that
\begin{equation}
    loss_e = -\Sigma_{i=1}^{C_{gt}}y_i\log P_{e,i}
\end{equation}
, where $y_i$ is the ground truth and $P_{e,i}$ is the predicted probability of the class $i$.

The dropped embedding matrix $Y_{d}$ is used to suppress irrelevant features, thereby increasing the distinction between foreground and background. The loss term is constructed as the mean squared error between the predictions derived from $Y_{d}$ and a pseudo label of $-1$ 
\begin{equation}
    loss_{d} = \Sigma_{i=1}^{C_{gt}}(P_{d,i} + 1)^2
\end{equation}
, where $P_{d,i}=\tanh{Y_d}$.
The total loss of the BS module is the sum of classification loss ($loss_e$) and the background suppression loss ($loss_d$):
\begin{equation}
    loss_{bs} = \lambda_e loss_e+ \lambda_d loss_d
\end{equation}
, where $\lambda_e$ and $\lambda_d$ are hyperparameters defined empirically.
\subsubsection{High-temperature Refinement}
Multi-scale context information has proven to be essential for object detection tasks \cite{wang2018multi}.
Inspired by \cite{lin2019zigzagnet}, 
we establish connections between the feature maps of the same level in both top-bottom and bottom-top pyramids to facilitate the exchange of multi-scale context information.
This information exchange enables the model to discern broad areas and acquire diverse representations in the earlier layers, while the later layers concentrate on finer and discriminative features.
The loss function computes the Kullback-Leibler divergence between the feature map on the bottom-top branch and the feature map on the top-bottom branch, both on the same scale. 
\begin{equation}
    P_{i1}=\text{LogSoftmax}(Y_{i1}/T_e)
\end{equation}
\begin{equation}
    P_{i2}=\text{Softmax}(Y_{i2}/T_e)
\end{equation}
\begin{equation}
    loss_r = P_{i2}\log \frac{P_{i2}}{P_{i1}}
\end{equation}
, where $T_e$ is the refinement temperature at training epoch $e$, decreasing as the training goes, followed by the function:
\begin{equation}
    T_e = 0.5^{\frac{e}{-\log_2(0.0625/T)}}
\end{equation}
The initial temperature $T$ is set to a high value, following the observations in \cite{hinton2015distilling}.
A higher temperature encourages the model to explore various features, even if the predictions are less accurate.
As training advances, the temperature gradually decreases, prompting the model to concentrate more on the target class and acquire more discriminative features. This decay policy enables the model to attain diverse and refined representations, facilitating accurate predictions.

The total loss of the back-end model is:
\begin{equation}
    L_{back-end} = loss_{bs} + loss_r
\end{equation}

\section{Experiments}
\label{section: experiments}
\subsection{Dataset and Implementation Details}
\label{subsection: dataset}
The datasets we used are 
SeaShips \cite{shao2018seaships}, and Singapore Maritime Dataset - Plus (SMD-Plus) \cite{kim2022object}.
The SMD-Plus is built upon the SMD dataset \cite{lee2018image} with corrected labels and improved bounding box annotations for small maritime objects. 
SeaShips consists of over 31455 images of ships with annotations, comprising 6 ship types.

In the training phase, we first trained the localizer within the front-end model, with emphasis placed on minimizing the loss terms $l_{obj}$ and $l_{reg}$.
The dataset underwent augmentations including blur, grayscale conversion, brightness adjustment and contrast, color jitter, and random gamma transformations.
After the training of the localizer, we fixed its parameters and merged it with the classifier.
The forward pass of this amalgamated model generates region proposals which are treated like fixed, pre-computed proposals when training the classifier on the front-end.
The dedicated difficulty estimator was trained based on the results of the edge classifier.
Samples erroneously classified by the edge classifier were ascribed to the label ``hard''.

The back-end model was trained on these ``hard'' samples and all the objects cropped from the dataset based on their annotations.
They underwent a preprocessing stage where they were resized and augmented including random horizontal flip, blur, and normalization.
During training, the learning rate was set to 0.0005, with cosine decay and weight decay set to 0.0005. 
Stochastic Gradient Descent (SGD) was utilized as the optimizer, utilizing a batch size of 8, and implementing gradient accumulation steps set to 4.
The selection of K in BS adheres to the principle that the condition $K_i > K_j$ holds when $i < j$. Specifically, values for $K_1$, $K_2$, $K_3$, and $K_4$ are set to 256, 128, 64, and 32, respectively.
The initial temperature $T$ was set to 128.
All models are trained on a server with a single Nvidia GeForce RTX 4090, utilizing PyTorch as the primary implementation framework.


\begin{table}[]
\centering
\caption{Performance Comparisons between Our Framework and Baselines on Different Datasets\\
\footnotesize{Data Transmission Volume Ratio (DTVR) is the ratio of the data volume transmitted to the cloud of our framework and that of a centralized manner.\\
Energy Consumption Ratio (ECR) is the ratio of the energy consumption of our framework and that of a centralized manner.}
}
\label{table: performance}
\resizebox{\columnwidth}{!}{%
\begin{tabular}{c|cccc}
\toprule
Dataset  & Baseline(mAP@50) & Ours(mAP@50) & TDVR & ECR \\ \midrule
SeaShips &    0.879\cite{liu2021enhanced}  
&    0.922          &    4.57\%                           &    27.3\%                      \\ 
SMD-Plus      &    0.898 \cite{kim2022object}               &    0.917          &      7.17\%                         &    31.6\%                      \\ \bottomrule
\end{tabular}
}
\end{table}

\begin{figure}[!t]%
\centering
\includegraphics[width=1\columnwidth]{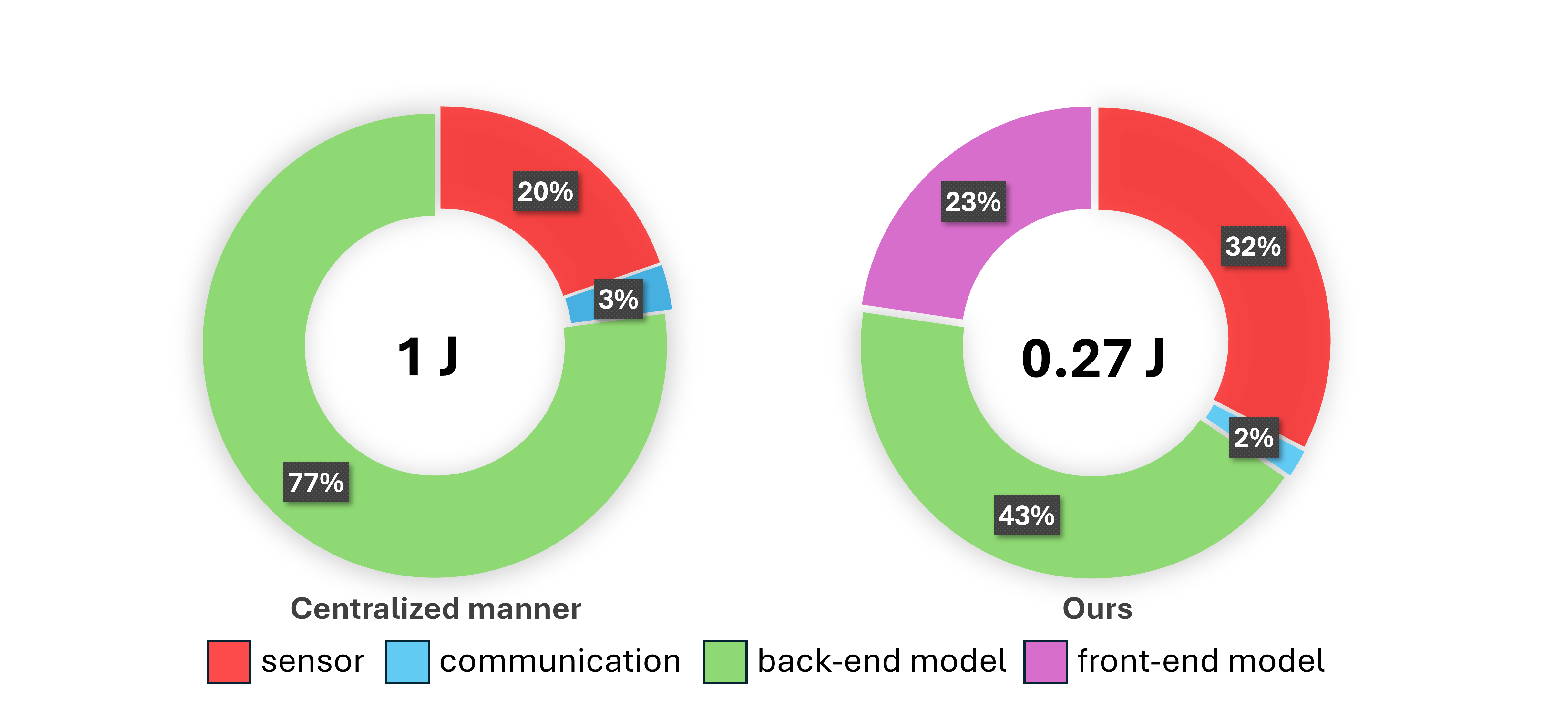}
\captionsetup{skip=0.05\baselineskip} 
\caption{Normalized energy consumption breakdowns of the conventional centralized manner and our system.}
\label{fig: energy breakdown}
\end{figure}

\begin{figure}[t]%
\centering
\includegraphics[width=1\columnwidth]{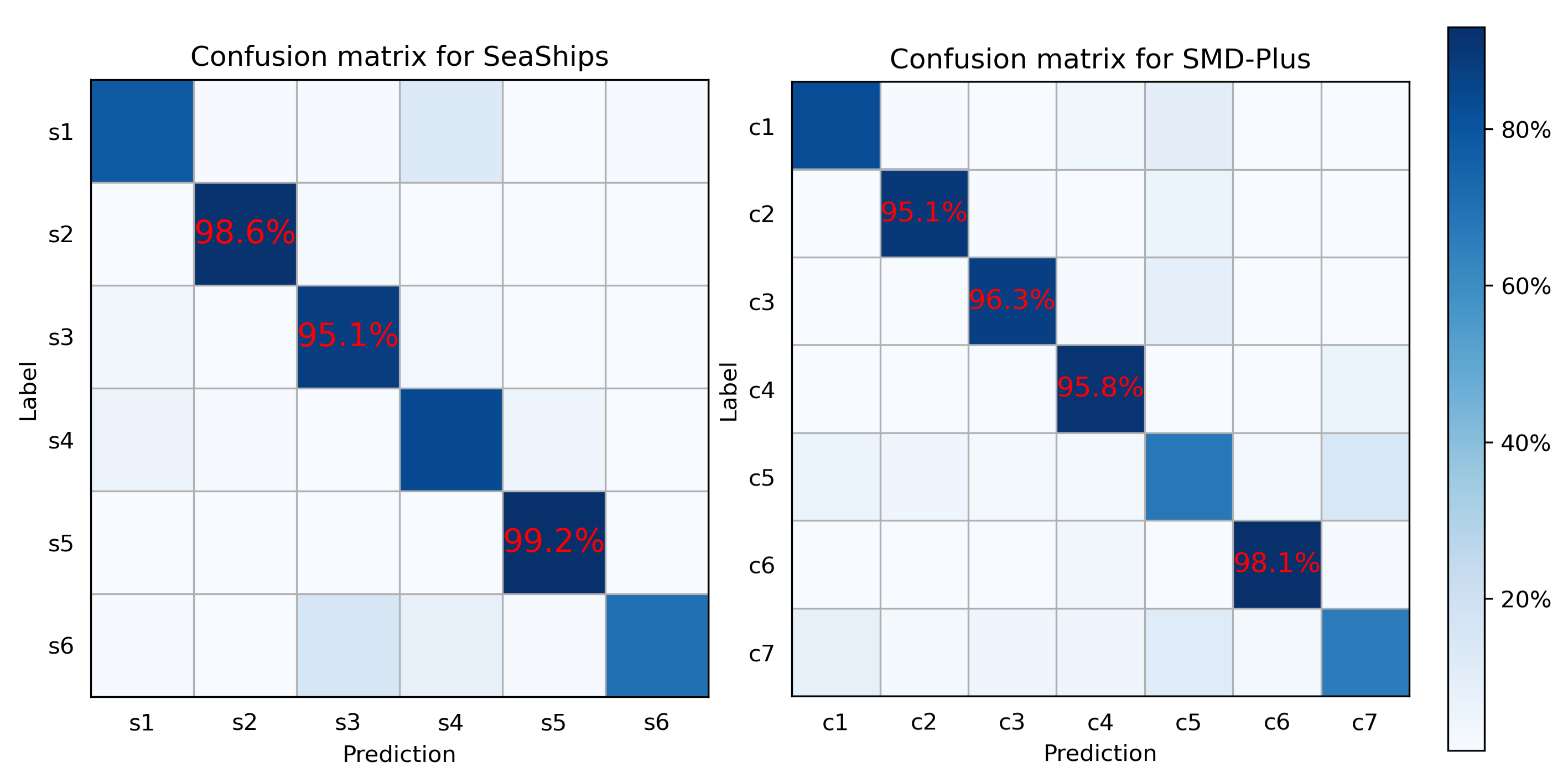}
\captionsetup{skip=0.05\baselineskip} 
\caption{
\textbf{Confusion Matrices of our back-end model}.\\
s1: bulk cargo carrier, s2: container ship, s3: fishing boat, s4: general cargo ship, s5: ore carrier, s6: passenger ship,
c1: ferry, c2: buoy, c3:
vessel ship, c4: boat, c5: kayak, c6: sail boat, c7: others.
}
\label{fig: confusion_matrix}
\end{figure}

\begin{table*}[]
\caption{Latency \& Power Measurement across Platforms \\
\footnotesize{* marks utilizing R Pi as the host machine}}
\label{table: power and latency}
\resizebox{\textwidth}{!}{%
\begin{tabular}{c|cccccccc|cc}
\hline
\multicolumn{1}{l|}{} &
  \multicolumn{8}{c|}{\cellcolor[HTML]{FFFC9E}Front-end Model on Edge Device} &
  \multicolumn{2}{c}{\cellcolor[HTML]{9AFF99}Back-end Model on Cloud Platform} \\ \hline
Platform     & Orin & Orin Nano & \multicolumn{1}{c|}{Nano} & TPU USB & TPU Dev & \multicolumn{1}{c|}{TPU Mini} & ZCU104        & Kria 260        & RTX 4090 & Alveo U200 \\ \hline
Host &
  Cortex-A78AE &
  \multicolumn{2}{c|}{Cortex-A57} &
  Cortex-A72* &
  Cortex-A53 &
  \multicolumn{1}{c|}{Cortex-A35} &
  \multicolumn{2}{c|}{Cortex-A53} &
  AMD Ryzen 5955WX &
  Intel i9-12900 \\
Kernel       & \multicolumn{3}{c|}{GPU}                     & \multicolumn{3}{c|}{Edge TPU Coprocessor}         & \multicolumn{2}{c|}{Xilinx DPU} & GPU      & Xilinx DPU \\ \hline
Framework    & \multicolumn{3}{c|}{PyTorch}                 & \multicolumn{3}{c|}{TensorFlow Lite and PyCoral}  & \multicolumn{2}{c|}{Vitis AI}   & PyTorch  & Vitis AI   \\ \hline
Latency (ms) & 77   & 86        & \multicolumn{1}{c|}{350}  & 15.9    & 9       & \multicolumn{1}{c|}{58.1}     & 142           & 126.35          & 24.53    & 16.8       \\
Power (W)    & 22.9 & 7.3       & \multicolumn{1}{c|}{3.9}  & 5.02    & 3.47    & \multicolumn{1}{c|}{0.92}     & 8.9           & 7.6             & 245      & 17.8       \\ \hline
\end{tabular}
}
\end{table*}

\subsection{Evaluations}
The comparisons of our model and the state-of-the-art (SOTA) on different datasets are shown in TABLE~\ref{table: performance}.
The mAP@0.5 for our framework surpasses that of the SOTA by 4.3\% on SeaShips and by 1.9\% on SMD-Plus datasets.
Regarding system efficiency, our framework reduces data transmission volume (DTV) by 95.43\% on SeaShips and 92.83\% on SMD-Plus. Additionally, our framework consumes only 27.3\% and 31.6\% of the energy compared to the centralized approach. 
For deployment, we utilized the Google TPU Dev Board for our front-end model and the FPGA Xilinx Alveo U200 for our back-end model. 
Further discussion on hardware settings and power consumption is provided below.
The normalized energy consumption breakdowns are shown in Fig.~\ref{fig: energy breakdown}. 
Despite our system incurring additional energy consumption for the front-end model compared to the conventional centralized approach, it executes most tasks efficiently, resulting in substantial savings in both communication and inference costs on the cloud.
The confusion matrices of the back-end model on the datasets are illustrated in Fig.~\ref{fig: confusion_matrix}, demonstrating its superior performance in tiny object fine-grained classification. 
In SeaShips, three out of six classes attain a test accuracy exceeding 95\%, whereas, in SMD-Plus, four out of seven classes achieve a test accuracy surpassing 95\%.

We deployed our front-end model on various edge devices and our back-end model on a set of cloud settings. 
The measured inference speed and power consumption are recorded in TABLE.~\ref{table: power and latency}. 
For edge devices, we explored three distinct hardware platforms: \textbf{edge GPU}, \textbf{edge TPU}, and \textbf{edge FPGA}.
Specifically, for edge GPU, we assessed the Nvidia \textit{Jetson Orin} (Orin), \textit{Jetson Orin Nano} (Orin Nano), and \textit{Jetson Nano} (Nano). 
Regarding edge TPU, we evaluated the Google \textit{TPU USB}, \textit{TPU Dev Board} (TPU Dev), and \textit{TPU Dev Board Mini} (TPU Mini), utilizing R Pi as the host machine for TPU USB (marked $\ast$ in TABLE.~\ref{table: power and latency}), while the other TPU development boards possess host CPUs \cite{ni2022online}. 
For edge FPGA, our investigation included the \textit{Xilinx ZCU104} and \textit{Kria 260}. 
The model kernel for R Pi and Nvidia GPU was implemented using PyTorch. 
For edge TPU, TensorFlow Lite and PyCoral API were employed for quantization and deployment. 
As for Xilinx FPGA, Vitis AI was utilized to map the model onto the Xilinx deep processing unit (DPU). 
Regarding host machines, we explored both CPU-GPU and CPU-FPGA heterogeneous platforms. 
For CPU-GPU heterogeneous platforms, the host CPU utilized was the AMD Threadripper, while the Nvidia RTX 4090 was evaluated for kernel GPU. 
For CPU-FPGA heterogeneous platforms, the host CPU selected was the Intel i9-12900, and the kernel FPGA was the Xilinx Alveo U200.

To meet the real-time requirements, the latency of the front-end devices should be smaller than 66ms \cite{iancu2021aboships}.
Considering the constraint, we believe edge TPU families show more advanced performance than edge GPUs and FPGAs since only they make inferences in real time while consuming few power (less than 5W), which is ideal for low-power settings. Our model's performance on the NVIDIA Jetson series is constrained by PyTorch's optimization for embedded GPUs, in comparison to TPU. The FPGA DPU, on the other hand, has lower frequency (only 300 MHz) and limited host-kernel bandwidth when compared to TPU platforms~\cite{lee2023comprehensive}. To achive better performance on FPGA, customized data path IP and computing unit IP are necessary~\cite{chen2023hypergraf}. For back-end model, both settings satisfy the real-time requirements.
Xilinx Alveo U200 is faster and more power-efficient than Nvidia RTX 4090.

\begin{figure}[!t]%
\centering
\includegraphics[width=1\columnwidth]{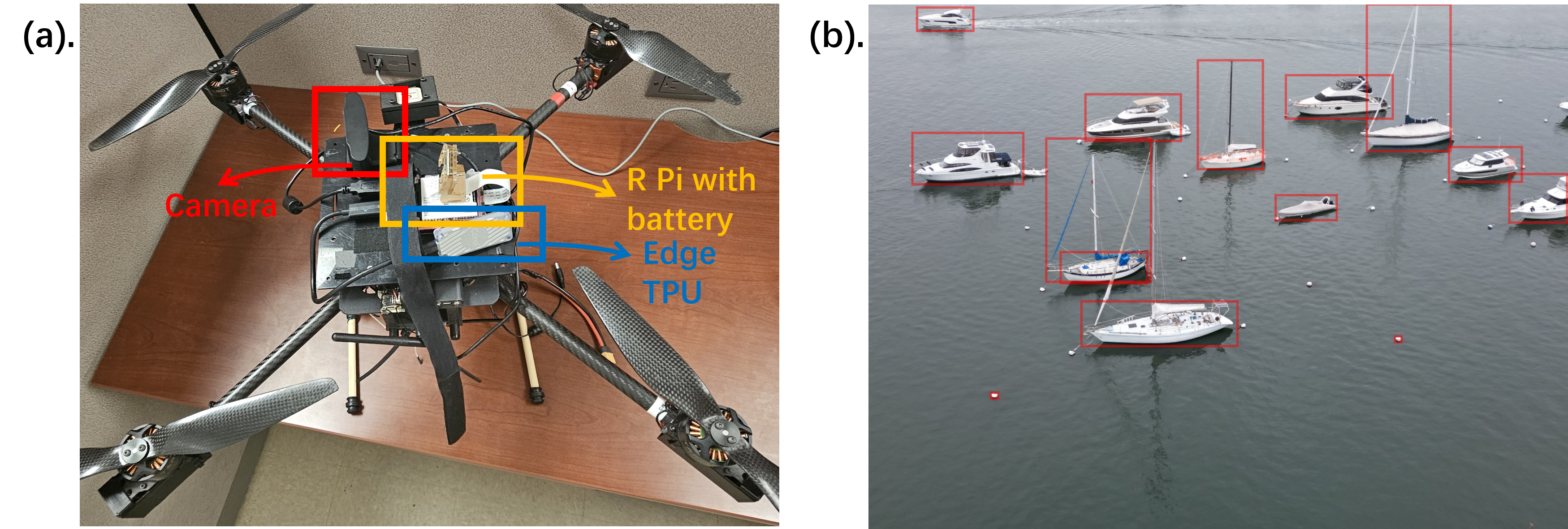}
\caption{\textbf{Real scenario deployment}. (a). Drone setup. 
(b). An example of the inshore picture captured by the drone with detection results. Red rectangles outline the detected objects.}
\label{fig: drone}
\end{figure}

We deployed the front-end model onto the drone and conducted testing in a real-world scenario. Fig.~\ref{fig: drone}(a) illustrates the drone outfitted with our front-end system, while (b) displays an example of the detection outcome generated by our system. 
The drone we used is Holybro PX4. The edge device we chose was a TPU USB with R Pi as a host.
The ships appearing in the picture are all detected and cropped precisely.
However, there are two buoys are also detected as ships.
We assume this is because they show a relatively similar shape to ships from the bird's view. 
Since the dataset we used for training lacked the data from the bird's view, we are optimistic that the performance of our model can be boosted when we collect more data from the same angle.

\section{Conclusion}

Our study introduces an edge-cloud collaborative sensing system specifically designed to address the multifaceted challenges of marine object detection. 
The system architecture consists of a low-power device-tailored front-end model for initial processing, and a transformer-graph convolutional network-based back-end model for fine-grained classification.
This dual-model framework enhances both performance and efficiency. 
Experiments are conducted across a spectrum of embedded system platforms, coupled with the validation of the system's effectiveness on a drone.

\section*{Acknowledgements}
This work was supported in part by the DARPA Young Faculty Award, the National Science Foundation (NSF) under Grants \#2127780, \#2319198, \#2321840, \#2312517, and \#2235472, the Semiconductor Research Corporation (SRC), the Office of Naval Research through the Young Investigator Program Award, and Grants \#N00014-21-1-2225 and N00014-22-1-2067. Additionally, support was provided by the Air Force Office of Scientific Research under Award \#FA9550-22-1-0253, along with generous gifts from Xilinx and Cisco.

\bibliographystyle{ACM-Reference-Format}
\bibliography{sample-base}

\end{document}